\documentclass[10pt,twocolumn,letterpaper]{article}
\pdfoutput=1
\usepackage[pagenumbers]{wacv}

\usepackage{graphicx}
\usepackage{amsmath}
\usepackage{amssymb}
\usepackage{booktabs}

\usepackage{tikz} 
\usepackage{pgfplots} 

\usepackage{multirow,longtable} 
\usepackage{tabularx} 
\usepackage{colortbl}
\usepackage{caption}
\usepackage{placeins}
\usepackage{notoccite}

\usepackage[edges]{forest}

\definecolor{folderbg}{RGB}{124,166,198}
\definecolor{folderborder}{RGB}{110,144,169}
\definecolor{IGNGREEN}{RGB}{153, 211, 142}
\definecolor{TITLES}{RGB}{153, 211, 142}
\definecolor{TITLES_PRE}{RGB}{247, 212, 188}
\newlength\Size
\tikzset{%
  folder/.pic={%
    \filldraw [draw=folderborder, top color=folderbg!50, bottom color=folderbg] (-1.05*\Size,0.2\Size+5pt) rectangle ++(.75*\Size,-0.2\Size-5pt);
    \filldraw [draw=folderborder, top color=folderbg!50, bottom color=folderbg] (-1.15*\Size,-\Size) rectangle (1.15*\Size,\Size);},
  file/.pic={%
    \filldraw [draw=folderborder, top color=folderbg!5, bottom color=folderbg!10] (-\Size,.4*\Size+5pt) coordinate (a) |- (\Size,-1.2*\Size) coordinate (b) -- ++(0,1.6*\Size) coordinate (c) -- ++(-5pt,5pt) coordinate (d) -- cycle (d) |- (c) ;},
}
\forestset{%
  declare autowrapped toks={pic me}{},
  pic dir tree/.style={%
    for tree={folder,font=\ttfamily,grow'=0,},
    before typesetting nodes={%
      for tree={edge label+/.option={pic me},},},
  },
  pic me set/.code n args=2{%
    \forestset{%
      #1/.style={inner xsep=2\Size,pic me={pic {#2}},}
    }
  },
  pic me set={directory}{folder},
  pic me set={file}{file},
}

\newenvironment{Tabular}[2][1]
  {\def\arraystretch{#1}\tabular{#2}}
  {\endtabular}
\usepackage[pagebackref,breaklinks,colorlinks]{hyperref}

\usepackage[capitalize]{cleveref}
\crefname{section}{Sec.}{Secs.}
\Crefname{section}{Section}{Sections}
\Crefname{table}{Table}{Tables}
\crefname{table}{Tab.}{Tabs.}

\begin{document}

\title{FRACTAL: An Ultra-Large-Scale Aerial Lidar Dataset for 3D~Semantic~Segmentation of Diverse Landscapes}
\author{{Charles Gaydon \hspace{1.5cm} Michel Daab \hspace{1.5cm} Floryne Roche}\\
{\normalsize Institut national de l’information géographique et forestière (IGN), France}\\
{\tt\small fractal.dataset@ign.fr}
}
\maketitle

\begin{abstract}
    Mapping agencies are increasingly adopting Aerial Lidar Scanning (ALS) as a new tool to map buildings and other above-ground structures. Processing ALS data at scale requires efficient point classification methods that perform well over highly diverse territories. Large annotated Lidar datasets are needed to evaluate these classification methods, however, current Lidar benchmarks have restricted scope and often cover a single urban area. To bridge this data gap, we introduce the FRench ALS Clouds from TArgeted Landscapes (FRACTAL) dataset: an ultra-large-scale aerial Lidar dataset made of 100,000 dense point clouds with high quality labels for 7 semantic classes and spanning 250 km². FRACTAL achieves high spatial and semantic diversity by explicitly sampling rare classes and challenging landscapes from five different regions of France. We describe the data collection, annotation, and curation process of the dataset. We provide baseline semantic segmentation results using a state of the art 3D point cloud classification model. FRACTAL aims to support the development of 3D deep learning approaches for large-scale land monitoring. 

\end{abstract}

\section{Introduction}\label{sec:intro}

High-density Aerial Lidar Scanning (ALS) is now recognized as a powerful remote sensing modality to support important public actions such as ecological monitoring \cite{wwf_lidar} and risk management (e.g., for floods \cite{lidar_flood_review} and forest fires \cite{fire_lidar_review} prevention). Detailed 3D mapping also contribute to the consolidation of existing geographic databases (e.g., by updating building footprints). The reduction in costs associated with the acquisition, storage, processing, and dissemination of Aerial Lidar Scanning (ALS) data has paved the way for its production at unprecedented scales. 

\begin{figure}[h!]
\centering
\includegraphics[width=0.46\textwidth]{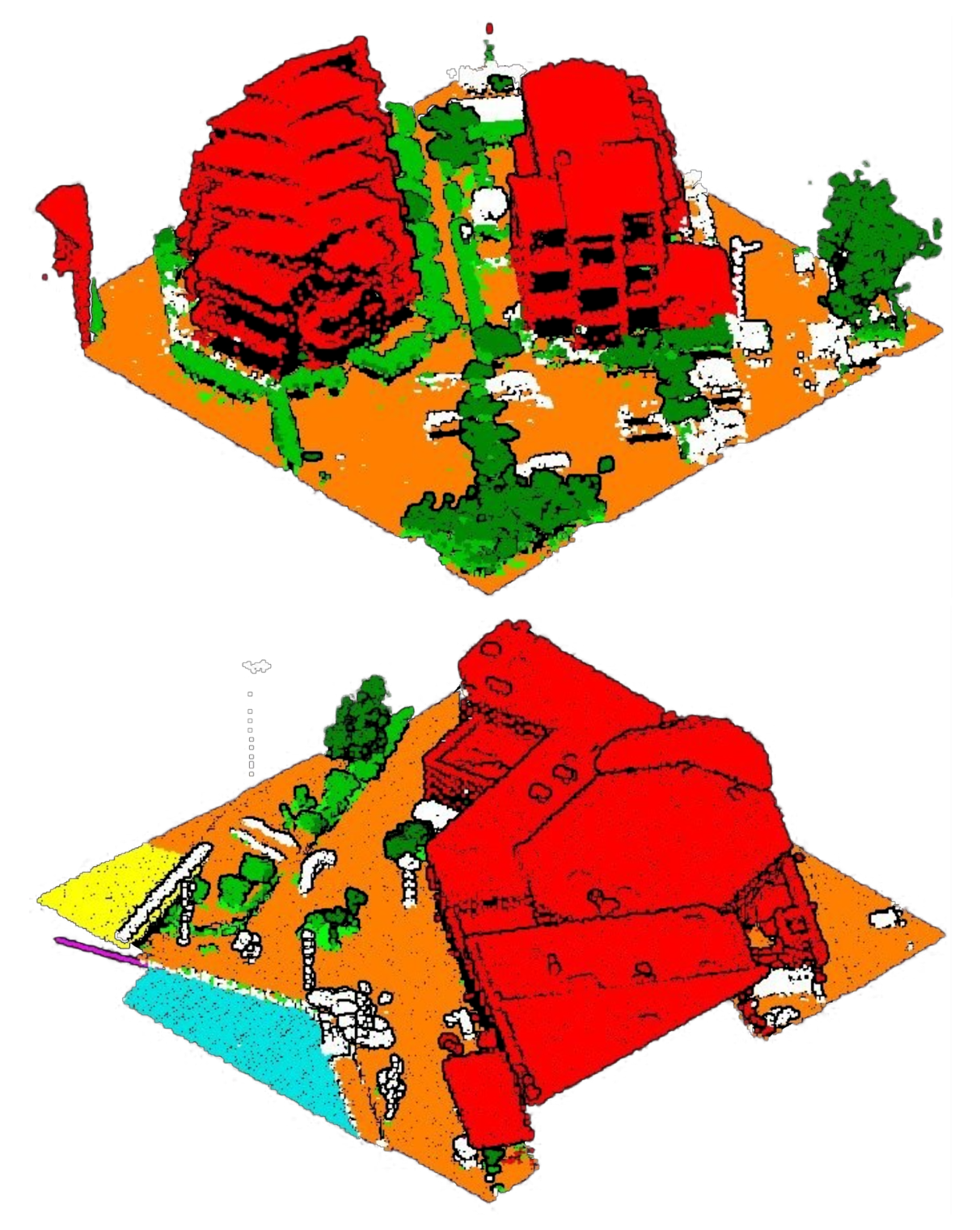}
\caption{Random scenes from our FRACTAL dataset. Each point is colored according to its semantic class; ground (orange), vegetation (greens), building (red), water (cyan), bridge (yellow), permanent structure (purple), other (white).}\label{fig:two_samples}
\end{figure}

In recent years, ALS has been increasingly used by public authorities at the regional and national levels. In a comprehensive report, \cite{kakoulaki} describe the availability of non-commercial Lidar data across Europe: at least a dozen European countries have already performed nationwide Lidar acquisitions, with densities from 10 to 40 pts/m². In France, the national mapping agency (\textit{Institut national de l'information géographique et forestière, IGN}) aims to map the entire French territory with high-density ALS point clouds (10 pulses/m², about 40 pts/m²) by 2026, in a program called Lidar HD (for "High Density") \cite{lidarhd}.

Point cloud classification is a prerequisite to most downstream use of Lidar products. For example, ground points must be identified to produce accurate Digital Terrain Models (DTMs) and separated from vegetation points to normalize tree heights in forest applications, e.g., for forest biomass estimation. Buildings are reconstructed from building points for urban modeling, while city trees are mapped from clusters of vegetation points to manage urban biodiversity.

The complexity, volume, and lack of structure of Lidar data make manual labelling of Lidar data time and labor-intensive; existing commercial software all require some form of manual input to reach satisfying results \cite{lidar_software_review}. Deep learning methods are increasingly available, especially since the publication of PointNet \cite{pointnet} in 2016, and of PointNet++ \cite{pointnet2} a year later.

Researchers need large annotated Lidar datasets to develop and evaluate their 3D deep learning methods. This need motivates us to release the FRench ALS Clouds from TArgeted Landscapes, or FRACTAL. FRACTAL is a large dataset for the semantic segmentation of ALS data, that spans 250 km². FRACTAL is made of 100,000 distinct point clouds, each spanning 50 $\times$ 50 m and typically containing 20k to 40k labelled points (37 pts/m² on average).

The dataset is built from the French Lidar HD data \cite{lidarhd}, using a sampling scheme that explicitly concentrates rare classes, rare objects, and challenging landscapes. Being sampled from an initial area of 17,440 km² in 5 French regions, it is characterized by a high variety of scenes. 

Point clouds are labelled with seven common semantic classes: ground, vegetation, building, water, bridge, permanent structure, and other. The labels were produced with automated processes and then verified and corrected by Lidar operators to achieve high quality classification. FRACTAL is overall the largest open benchmark dataset for 3D semantic segmentation, and the first to offer the diversity of landscapes inherent to large-scale land monitoring.

We make the following contributions:

\begin{itemize}
\item Define a general framework to catalog and sample diverse point clouds from large open ALS archives.
\item Introduce FRACTAL: a benchmark dataset for 3D point cloud semantic segmentation, sampled from verified areas of the French Lidar HD data. The sampling has explicit consideration for spatial diversity and landscape diversity, making FRACTAL suitable for the evaluation of point cloud classification methods against the specific challenges of land monitoring.
\item Set a baseline evaluation of segmentation performance on the dataset using a state-of-the-art deep point cloud segmentation model, to show its potential for benchmarking.
\end{itemize}

Section \ref{related_work} lays out current ALS benchmarks and their shortcomings. Section \ref{section:data_sources} introduces the data sources used to create FRACTAL. Section \ref{section:fractal_the_dataset} presents the strategy for cataloguing and sampling data patches at scale, and highlights the high diversity of the resulting dataset. Section \ref{section:experimental_results} describes the first experimental results on the benchmark.

\section{Related work}\label{related_work}
ALS differs from ground-based Lidar in critical ways: airborne Lidar has nadir orientation instead of lateral orientation, and a lower but more homogeneous point density \cite{dales_dataset}. Most importantly, airborne Lidar can be used to collect data at a much larger scale than ground-based Lidar, which adds specific challenges to their classification. Vast territories the size of a country have diverse landscapes and vegetation, along with unique signs of human activity, leading to high intraclass heterogeneity. Additionally, geographic data are characterized by spatial autocorrelation and a long-tail distribution, resulting in a variety of rare scenes that may be spatially concentrated, i.e., globally rare but locally frequent, like wind turbines or greenhouses.

The specific nature and scope of ALS must be addressed in the evaluation of 3D point cloud semantic segmentation methods. Having large representative annotated datasets is critical, however, current benchmark datasets come up short: limited in size and spatial diversity, they cannot attest to the capacity of 3D deep learning models to deal with the challenges of large-scale land monitoring. In a comprehensive review of current Lidar benchmark datasets for 3D point cloud semantic segmentation, authors found 26 datasets \cite{cenagis_als_dataset}; only six of them have ALS data, which we report in \cref{tab:datasets} along with their proposed dataset (CENAGIS-ALS) and ours (FRACTAL).

\begin{table*}
\begin{minipage}{\textwidth}
\centering
\setlength{\tabcolsep}{7.5pt}
\renewcommand{\arraystretch}{1.}
\caption{Benchmark datasets for 3D semantic segmentation of ALS data. We include the datasets listed by \cite{cenagis_als_dataset} along with their proposed benchmark dataset CENAGIS-ALS, and our own dataset FRACTAL. ALS: Aerial Lidar Scanning; ULS: Unmanned Lidar Scanning.}
\begin{tabular}{lccrcrr}
\textbf{Reference} & \textbf{Year} & \textbf{Sensor Platform} & \textbf{Area (km²)} & \textbf{Classes} & \textbf{Points} & \textbf{Density (pts/m²)}  \\ \hline
Vaihingen (ISPRS) \cite{vaihingen_isprs_dataset} & 2012 & ALS & 0.1 & 9 & 1.16M & 4-7  \\
DublinCity \cite{dublincity_dataset} & 2015 & ALS & 2 & 9 $\in$ 7 $\in$ 4 & 260M & 348  \\
LASDU \cite{lasdu_dataset} & 2020 & ALS & 1.02 & 5 & 3.12M & 3-4  \\
DALES \cite{dales_dataset} & 2020 & ALS & 10 & 8 & 500M & 50  \\
Hessigheim 3D \cite{h3d_dataset} & 2021 & ULS & 0.19 & 11 & 74M & 800  \\
OpenGF \cite{opengf_dataset} & 2021 & ALS & 47.7 & 2 & 542M & 11  \\
CENAGIS-ALS \cite{cenagis_als_dataset} & 2023 & ALS & 2 & 49 $\in$ 28 $\in$ 7 & 550M & 275  \\
\rowcolor[HTML]{e2e7f9} \textbf{FRACTAL (Ours)} & \textbf{2024} & \textbf{ALS} & \textbf{250} & \textbf{7} & \textbf{9261M} & \textbf{37}  \\
\end{tabular}

\label{tab:datasets}
\end{minipage}
\end{table*}

Five of these benchmarks cover only a tiny area, not exceeding 2 km².
This reduced scale contrasts with the larger support that other modalities enjoy in land monitoring, e.g., with massive land cover mapping benchmarks such as LandCover.ai \cite{landcoverai_dataset}, which covers 216 km² with VHR aerial images, and FLAIR \cite{flair_dataset}, which spans 817 km² across 50 distinct spatial domains with both VHR aerial images and Sentinel 2 time series. 

Furthermore, point densities in these ALS benchmarks is either too low for typical urban applications (3-4 pts/m² in LASDU \cite{lasdu_dataset}, 4-7 pts/m² in ISPRS Vaihingen \cite{vaihingen_isprs_dataset}) or unreasonably high (275 pts/m² in CENAGIS-ALS \cite{cenagis_als_dataset}, 348 pts/m² in DublinCity \cite{dublincity_dataset}, 800 pts/m² in Hessigheim 3D \cite{h3d_dataset}).

The Dayton Annotated Lidar Earth Scan (DALES) \cite{dales_dataset} attempts to provide a level of detail more representative of real-world use cases, with 8 semantic classes and a point density of 50 pts/m². It is a significant improvement in scale compared to previous urban ALS benchmarks: covering an area of 10 km², DALES is the largest dataset of its kind.

Unfortunately, all ALS benchmarks mentioned so far, including DALES, are limited to single urban areas. This results in adjacent training and test data, which is likely to result in overestimated performance metrics given the high autocorrelation of geographic data. Even with reduced ambition (e.g., only considering urban contexts), it is impossible to draw conclusions about the spatial generalizability of the benchmarked methods with current ALS benchmarks. In addition, considering only a single, small urban area hides the complexity of regional scale land monitoring. Overall, researchers need datasets with better territorial representativeness.

OpenGF \cite{opengf_dataset} is the one ALS benchmark that addresses these shortcomings. It is built from open large-scale ALS archives from 4 countries in 3 continents, via a careful selection of areas with high-quality point annotations in diverse landscapes (i.e., metropolis, small city, village, mountain). OpenGF is largest ALS benchmark to date, with 542M points over 50 km². Authors demonstrate the benefits of leveraging open Lidar assets to quickly create high-quality, diverse, large-scale benchmark datasets for 3D semantic classification. However, OpenGF is tailored for the specific task of ground/non-ground classification and was stripped of the semantic classes used in most applications (e.g., building, vegetation).

In summary, current ALS benchmarks for 3D point cloud segmentation are either limited in both volume and representativeness, and unsuitable to evaluate generalizability of methods, or are large and diverse enough but lack semantic depth (OpenGF). Our proposed dataset, FRACTAL, aims to address these shortcomings with: volume (250 km²), territorial representativeness (sampling from 17,440 km² in 5 distinct spatial domains), reduced influence of spatial autocorrelation (large contiguous test areas), and a semantic depth matching typical land monitoring applications (7 classes).

\section{Data Collection}\label{section:data_sources}
We achieve efficient, large scale dataset creation by leveraging an open ALS data archive: the Lidar HD data. To reflect real-world Lidar processing practices, clouds are colorized from national Very High Resolution (VHR) aerial imagery from the ORTHO HR database.

\subsection{ALS point clouds}\label{section:lidarhd}

The Lidar HD program \cite{lidarhd} is a national initiative that aims to provide a 3D description of the French territory by 2026, using high-density ALS (10 pulses/m² or about 40 pts/m²). The data produced as part of this program (i.e., point clouds, Digital Terrain Models, Digital Surface Models) are made available as open data with extensive documentation \cite{lidarhd_documentation}. 

The program covers mainland France and its overseas territories for a total of 550,000 km². It consists of four phases: data acquisition, storing, processing, and dissemination, along user support. The data acquisition and processing are sequenced in blocks of 50 $\times$ 50 km and must be compatible with a variety of Lidar sensors (Leica, Riegl, and Teledyne/Optech), acquisition seasons (leaf-on or leaf-off), and landscapes (e.g., urban, rural, mountains, seashores, overseas territories). The point clouds are disseminated with semantic segmentation labels from 11 classes: unclassified, ground, vegetation (low, medium, and high), buildings, water, bridge deck, permanent structures, artifact, synthetic. The specification of this nomenclature is detailed in \cref{appendix:lhd_nomenclature}.

The classification of Lidar HD data comes in two flavors:

\textit{Classified Lidar HD} ~ Results from a fully automated classification process using commercial software and deep learning models.

\textit{Optimized Lidar HD} ~ Results from automatic classification followed by manual corrections. The annotations are then audited via visual inspection for approximately 10\% of the delivered point clouds. All classification errors are listed and rated for severity, with particular attention to confusions in buildings, bridges and ground. Special importance is given to the exhaustivity of individual buildings, with a minimal recall of 99.9\% in high-stakes areas (e.g., flood-prone, urban) and a minimal recall of 99.5\% elsewhere. The classification is validated if it meets the requirements for real-world use; otherwise, it undergoes further manual corrections until it reaches the desired quality. 

\subsection{VHR aerial images}\label{sec:vhr}

The ORTHO~HR® \cite{bdortho} is a mosaic of VHR aerial images acquired during national surveys. The individual images are mapped onto a cartographic coordinate reference system and projected on the RGE ALTI Digital Terrain Model for orthorectification. ORTHO~HR images have a high spatial resolution of 0.20 m, and near infrared, red, green, and blue channels. Radiometric processing methods, including equalization and global correction, are applied to obtain the final product. 

Aerial surveys are not synchronized with Lidar HD acquisitions and a variable time lag of up to 3 years might separate them. Indeed, the aerial images are renewed every three years, whilst Lidar data are processed as soon as they are acquired. In the interim, buildings may have been constructed and vegetation may have expanded or been cleared. Moving elements such as vehicles are unlikely to be consistent. Also, the appearance of cultivated fields is likely to be inconsistent between modalities due to different seasons of acquisition. Despite these discrepancies, working with colorized point clouds is recommended as it systematically yields better segmentation models. For all practical purposes, we document the year of acquisition of images and point clouds as part of the dataset's metadata.

\section{FRACTAL: the dataset}\label{section:fractal_the_dataset}
\subsection{Area of interest}
\begin{figure}[h!]
\centering
\includegraphics[width=0.5\textwidth]{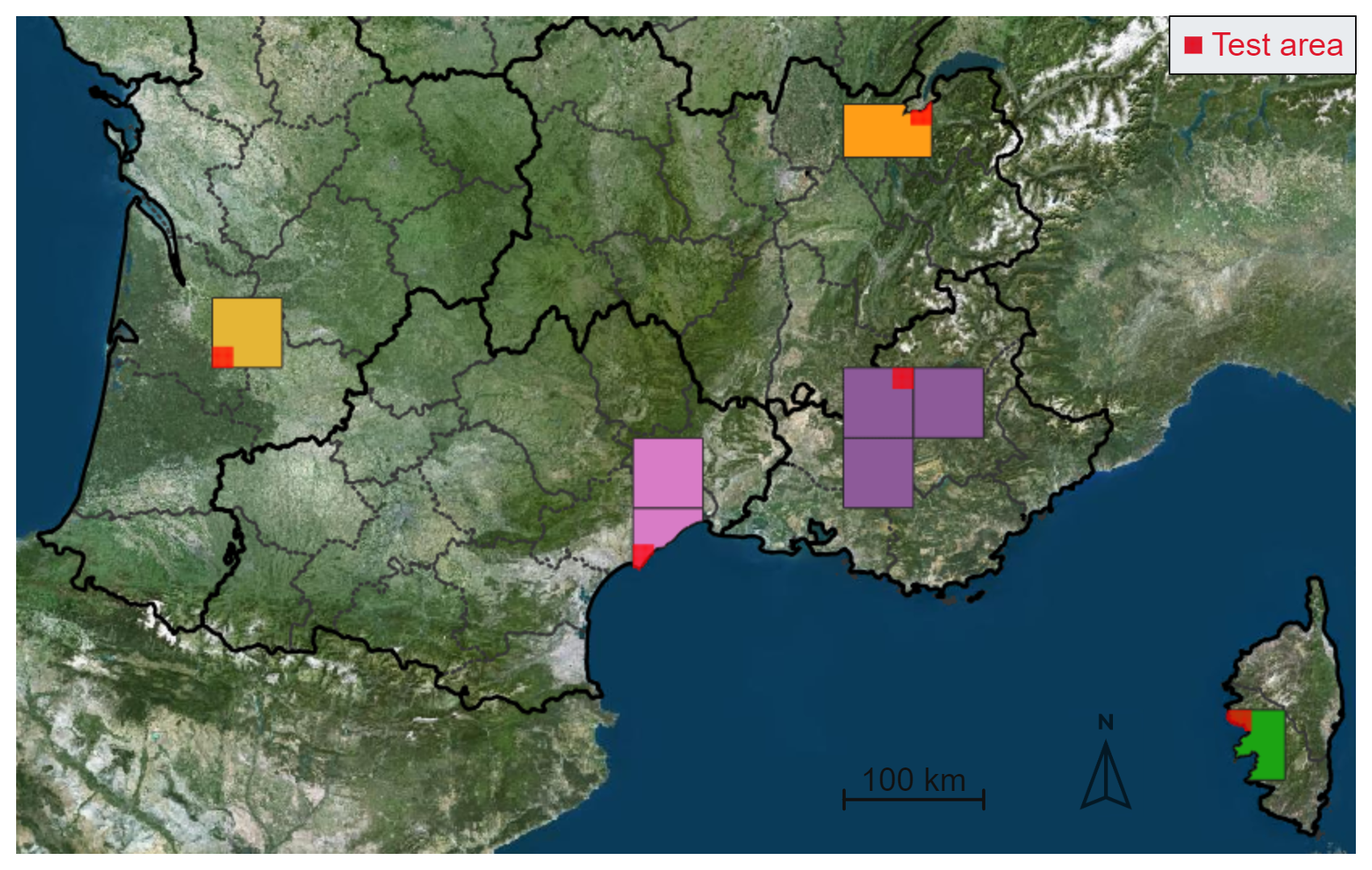}
\caption{The five spatial domains composing the 17,440 km² of the area of interest. The 1,049 km² reserved to sample the test set are highlighted in red.}
\label{fig:map}
\end{figure}

At the time of dataset creation, Lidar HD data is available for about half of metropolitan France, mainly in its southern half. To have the highest possible annotation quality, we restrict our sampling to Optimized Lidar HD i.e., to the areas that went through human verification and audit of the classification. \cref{fig:map} presents the 5 spatial domains we consider, each spanning 3456 km² on average, and at least 100 km distant from one another. The spatial domains compose an area of interest of 17,440 km² in Southern France, acquired with a variety of Lidar sensor (documented in Appendix \ref{appendix:sensors_subcontractors}).

To define a common setting for the benchmark of deep learning models, we reserve an area for testing in each spatial domain. Each of the 5 test areas span 210~km² on average, for a total of 1049~km². Having only a limited number of spatial domains with their own unique characteristics (seashores, mountains, field crops, etc.), test data and train data share the same spatial domains. Test areas are however contiguous, large, and distinct from the train areas. This enables spatial block validation and reduces the influence of spatial autocorrelation on model evaluation, which is critical when evaluating remote sensing solutions~\cite{spatial_autocorrelation}.

\subsection{Sampling methodology}\label{section:design_principles}

The sample unit is a 50 × 50 m square patch, which provides a compromise between detailed structures and long-range spatial dependencies. A square shape facilitates data extraction and visualization of data patches. Also, training models on square point clouds favors the use of non-overlapping tiling at inference time, thus gaining efficiency by limiting redundant predictions.

We adopt the following design principles for sampling:

\textbf{Class rebalancing}: We explicitly address class imbalance, which is often detrimental to model training, by oversampling rarer classes such as water, bridge, and permanent structures.

\textbf{Uncertainty sampling}: We aim to increase the representation of scenes known to be challenging for classification models by oversampling specific landscapes and hardscapes, such as mountainous areas, seashores, or complex urban scenes.

\textbf{Spatial sampling as a proxy for landscape diversity}: Given the spatial autocorrelation of geographic data, we aim for the broadest spatial distribution of patches. This maximizes diversity within each category of scenes, and globally.

Scenes of special interest need first to be identified as such to consider them for sampling.
We define 19 scene types, which we list in \cref{tab:descriptors}. Twelve of them are related to the semantic classification of the point cloud and to the presence of certain objects within a semantic class. 
The seven other characterize patches in terms of landscapes, using heuristic definitions.

\begin{table*}
\begin{minipage}{\textwidth}
\centering
\setlength{\tabcolsep}{8pt}
\renewcommand{\arraystretch}{1.}
\caption{Scene types and their target minimal proportion for targeted sampling.}
\begin{tabular}{lllcc}
\textbf{Motivation} & \textbf{Scene Type} & \textbf{Definition} & \multicolumn{2}{l}{\textbf{Target (\%)}}  \\ \hline
\arrayrulecolor{lightgray!60}
\multirow{8}{*}{Classes} & \texttt{BUILD} & building $\geq$ 500 pts & 8 & \multirow{3}{*}{10} \\
 & \texttt{BUILD\_GREENHOUSE} & greenhouse (BD TOPO) & 1 &   \\
 & \texttt{BUILD\_BIG} & non-residential building (BD TOPO) & 1 &   \\ \cline{2-5}
 & \texttt{BRIDGE} & bridge $\geq$ 50 pts & 5 & 5  \\ \cline{2-5}
 & \texttt{WATER} & eau $\geq$ 50 pts & 4 & \multirow{2}{*}{5}  \\
 & \texttt{WATER\_SURFACE} & water area (BD TOPO) \& eau $\geq$ 50 pts & 1 &  \\ \cline{2-5}
 & \texttt{PERMSTRUCT} & permanent structure $\geq$ 50 pts & 3 & \multirow{3}{*}{5}  \\
 & \texttt{PERMSTRUCT\_PYLON} & pylon (BD TOPO) \& permanent structure $\geq$ 50 pts & 1 &   \\
 & \texttt{PERMSTRUCT\_ANTENNA} & antenna (BD TOPO) \& permanent structure $\geq$ 50 pts & 1 &  \\ \cline{2-5}
 & \texttt{OTHER} & unclassified $\geq$ 250 pts & 3 & \multirow{3}{*}{5}  \\
 & \texttt{OTHER\_PARKING} & parking lot (BD TOPO) \& unclassified $\geq$ 250 pts & 1 &   \\
 & \texttt{OTHER\_HIGHWAY} & highway (BD TOPO) \& unclassified $\geq$ 400 pts & 1 &   \\ \hline
\multirow{6}{*}{Landscapes} & \texttt{FOREST} & high vegetation $\geq$ 90\% of points & 5 & \multirow{6}{*}{20}  \\
 & \texttt{HIGHSLOPE1} & 35 m $\leq$ elevation gain $<$ 45 m & 2 &   \\
 & \texttt{HIGHSLOPE2} & elevation gain $\geq$ 45 m & 2 &   \\
 & \texttt{MOUNTAIN} & elevation $\geq$ 1000 m & 4 &   \\
 & \texttt{WATER\_ONLY} & water $\geq$ 50 pts \& ground $=$ 0 pts & 1 &   \\
 & \texttt{SEASHORE} & -10 $\leq$ elevation $<$ 10 \& water $\geq$ 50 \& ground $\geq$ 100 pts & 1 &   \\
 & \texttt{URBAN} & building $\geq$ 25\% of points & 5 &   \\

\end{tabular}

 \label{tab:descriptors}
\end{minipage}
\end{table*}

~

We adopt a simple yet effective sampling scheme consisting of targeted sampling followed by completion sampling: 

\textbf{Targeted sampling}: For each scene type, we sample patches with stratification on 1~$\times$~1~km Lidar HD tiles, until we reach the target proportion for that scene type. Each sampling is performed independently, with replacement, to ensure that the spatial diversity within each group is optimal; duplicates are then dropped.

\textbf{Completion sampling}: The dataset is completed with stratification on 1~$\times$~1~km Lidar HD tiles, this time without consideration for scene type. This is to ensure that we include more ordinary scenes such as field crops and rural areas, which are the majority.

Based on our own experience with training semantic segmentation models on Lidar HD data, we set a target dataset size of 250 km², or 100,000 patches -- a 70-fold reduction compared to the initial area of interest. This manageable size is suitable for common research practices and computing infrastructures. 

For model benchmarking, we define a reference train/val/test split with an 80/10/10 ratio: 225 km² of data go to model training, of which 25 km² are reserved for in-training evaluation, and an additional 25 km² of data are sampled from test areas and kept for model evaluation. Train and test areas are sampled independently with the same target proportions of each scene type. Patches for in-training validation are sampled from the train areas, again with stratification on Lidar HD tiles. This strategy results in the widest possible spatial distribution for each scene type, and ensures that all landscapes are equally represented in the train, val, and test sets. See Appendix \ref{appendix:descriptors_trainvaltest} for the proportions of scene types in each set.

\subsection{Spatial and landscape diversity in FRACTAL}\label{section:results_sampling}

\cref{tab:fractal_size} presents key figures of the dataset. Despite being 70 times smaller than the area of interest, the dataset incorporates data from all 17,440 initial Lidar HD tiles, ensuring comprehensive representation of its landscapes. With 9,261M points in 100,000 patches scattered over five vast spatial domains, FRACTAL has unprecedented territorial representativity, spatial diversity, and volume.

\begin{table}[htpb]
\setlength{\tabcolsep}{4.9pt}
\renewcommand{\arraystretch}{1.}
\caption{Size, areas, and number of points in FRACTAL compared to the area of interest. $r$ = ratio.}
\begin{tabular}{l|ccc|c}
 & \textbf{Area of interest} & $\rightarrow$ & \textbf{FRACTAL} & $\mathbf{r}$ \\ \hline
\textbf{Tiles} & 17440 & $\rightarrow$ & 17440 & 1 \\ \arrayrulecolor{lightgray!60}\hline\arrayrulecolor{black}
\textbf{Area (km²) }& 17440 & $\rightarrow$ & 250 & \multirow{2}{2em}{~70} \\
\textbf{Patches} & 6976000 & $\rightarrow$ & 100000 &  \\ \arrayrulecolor{lightgray!60}\hline\arrayrulecolor{black}
\textbf{Points (M)} & 661998 & $\rightarrow$ & 9261 & 71
\label{tab:fractal_size}
\end{tabular}
\end{table}

\begin{figure*}
    \vspace{-0.2cm}
    \centering
    \includegraphics[width=1.0\textwidth]{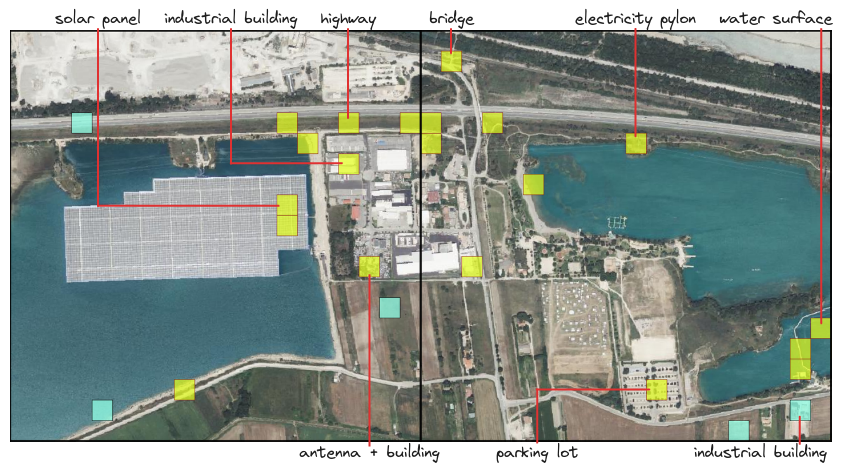}
    \vspace{-0.5cm}
    \caption{Enlarged view of two Lidar HD tiles and their sampled patches, most of which contain scenes of particular interest.}\label{fig:sampling_qgis_1x2}
\end{figure*}

The target proportions for each scene type are specified in \cref{tab:descriptors}. They are chosen so that about 20\% of sampled patches consist of specific landscapes, such as seashores and forests, while about 30\% of them contain specific structures and objects, such as highways and water surfaces. Half of the dataset is found by targeted sampling (47,253 patches, or 47.3\%), then completed (52,747 patches, or 52.7\%) until the target dataset size is reached. The target minimum proportion is achieved for all scene types except one: scenes with permanent structure points and with an antenna in the BD TOPO could not amount to 1\% in the train set (target of $n = 1000$~patches, 830 were found) and test set (target of  $n = 100$~patches, 65 were found).

\cref{fig:sampling_qgis_1x2} presents a visualization of the sampling in two Lidar HD tiles, where 25 patches are selected
from the 800 possible patches. Targeted sampling (yellow) over-concentrates specific landscapes such as water surfaces, specific classes such as buildings, and specific human-made structures such as parking lots and electricity pylons. On the other hand, four of the five patches selected by completion sampling (cyan) do not contain any predefined scene type. This illustrates the scarcity of complex scenes in geographical data and the need to explicitly target them to obtain a challenging and diverse dataset. See \cref{appendix:sampling_qgis_4x6} for a similar illustration at a larger scale.

\begin{figure*}[h!]
 \centering
    \vspace{-10pt}
    \includegraphics[width=1.0\textwidth]{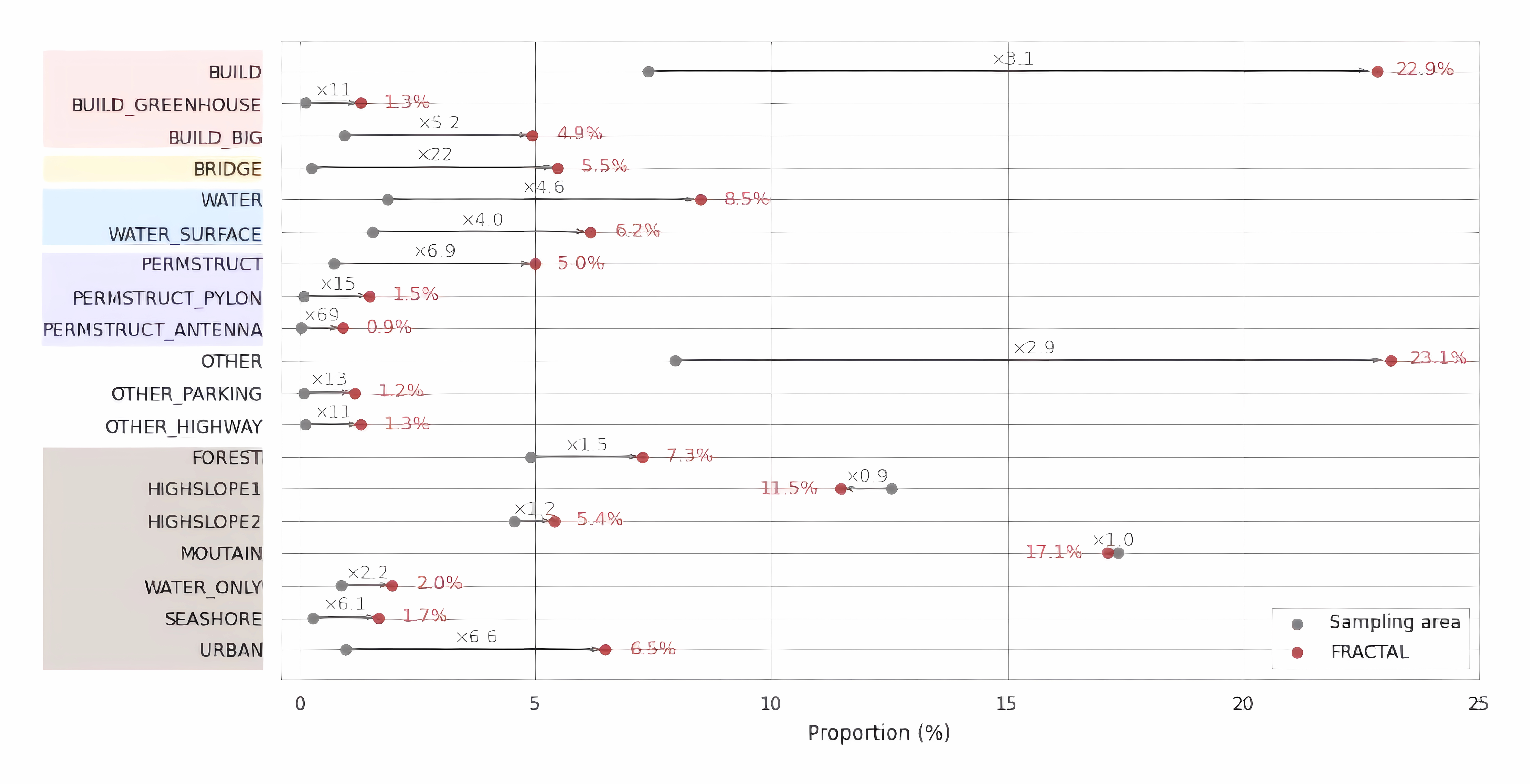}
    \vspace{-30pt}
    \caption{Proportional change of proportion of scene types in FRACTAL after sampling.}\label{fig:lollipop}
\end{figure*}

\cref{fig:lollipop} illustrates how our sampling increases the presence of targeted scene types in FRACTAL. The gain is most important for rarer classes, with up to a 22-fold concentration of scenes with a bridge. All scene types of interest end up with better representation in FRACTAL; except for high-slope scenes: most of them are spatially concentrated in mountainous areas, and thus slightly undersampled by completion sampling.  

Additional figures are reported in the appendices: proportions of each semantic class (\ref{appendix:all_sizes}) and prevalence of scene types in train, val, and test sets (\ref{appendix:descriptors_trainvaltest}).

\subsection{Data extraction and colorization}

From the sampled $50 \times 50$~m geometric patches, we extract the corresponding ALS point clouds. We colorize point clouds using near infrared, red, green and blue channels from the ORTHO HR images, based on the vertical alignment of points and pixels. Colorization is done regardless of potential obstructions: for instance, a ground point beneath a tree will be colored with the spectral information of the tree above it. 

Point clouds are made available in the LAZ 1.4 format defined by the American Society for Photogrammetry and Remote Sensing (ASPRS) \cite{asprs_las}. The following dimensions are relevant for semantic segmentation: xyz coordinates in Lambert 93 spatial reference system (EPSG:2154), intensity (16-bits encoding), return number and number of returns, scan angle, and near-infrared, red, green and blue (16-bits encoding). The Lidar HD classification nomenclature, detailed in Appendix \ref{appendix:lhd_nomenclature}, is left intact in the data patches.

Except for its colorization, point clouds in FRACTAL inherit the characteristics of the Lidar HD data, and we refer users to Section 2.2 of their official product description \cite{lidarhd_documentation} for more information.

\subsection{Limitations and recommendations}\label{section:limitations}

FRACTAL is sampled from an open ALS archive, and shares its limitations. The dataset only includes data that underwent human corrections and auditing (i.e. “Optimized Lidar HD”), however, the specification of the Lidar HD classification tolerates some errors. This tolerance is class specific; for instance, a perfect 100\% recall for individual buildings is not mandatory (as previously stated, see Section \ref{section:lidarhd}). 

Beyond errors, the very definition of “unclassified” points leads to potential confusions with other classes. This class is semantically ill-defined and may include points with blurred boundaries with adjacent classes such as building, ground, and vegetation. These imperfections are expected to have a negligible impact for model evaluation, but users should be aware that they exist when inspecting the data.

Point clouds are colorized with non-simultaneous aerial imagery (as explained in Section \ref{sec:vhr}), which can lead to color artifacts (typically for moving objects such as vehicles) and to misalignment between colors and shapes, which can blur object boundaries (e.g., roof points colored by a nearby tree). We observed only marginal impact on segmentation performances; overall, the colorization of the point clouds benefits segmentation models.

Using FRACTAL to train models for production should be done with caution. The dataset covers 5 spatial domains from 5 southern regions of metropolitan France. While large and diverse, it reprents only a fraction of the French territory and is not representative of its full diversity regarding either landscapes or human-made structures. Furthermore, domain shifts are frequent in aerial images due to different acquisition conditions and downstream data processing. When using models trained on the dataset to predict a classification in unseen regions, one should make the adequate verification and not assume its capacity to generalize. Similarly, considering new Lidar sensors should be done carefully. ALS data of comparable point densities (about 40 pts/m²) are expected to have consistent geometric characteristics, but users should still assess the model’s accuracy when predicting from alternative 3D data sources.

\section{Experimental results}\label{section:experimental_results}
\subsection{Task and metrics}\label{section:task_metrics}

FRACTAL is a benchmark dataset for semantic segmentation of ALS scenes into 7 semantic classes: other, ground, vegetation, building, water, bridge, and permanent structure. This reduced nomenclature is adapted from the Lidar HD nomenclature, which has 11 classes (see Appendix \ref{appendix:lhd_nomenclature}). Low, medium, and high vegetation are grouped together since they differ only in height above ground. Artifact and synthetic points are simply filtered out. The “unclassified” ASPRS class is renamed "other" for clarity. See \cref{appendix:nomenclatures} for a table of correspondence between the two nomenclatures.

We adopt the evaluation practices of similar Lidar datasets (e.g., DALES \cite{dales_dataset}) and use the mean Intersection over Union (mIoU) as our main metric. The "other" class is imperfect (see Section \ref{section:limitations}), but we decide not to exclude it from metric calculations since it contains well-defined objects that are relevant to scene segmentation (e.g., boats, containers). Considering the size of the dataset, imperfections of the "other" class are expected to have a negligible impact in model evaluation. 

For consistency with other benchmarks, we also report Overall Accuracy (OA). All metrics should be computed from the confusion matrix accumulated over all point clouds from the test set.

\subsection{Baseline model}

A baseline performance is established using Myria3D \cite{myria3d}. Myria3D is a 3D deep learning library developed at IGN, which leverages Pytorch-Lightning \cite{pytorch_lightning}, and the the 3D deep learning library PyTorch-Geometric \cite{pytorch_geometric}. It was explicitly built for the semantic segmentation of Lidar HD data. Scalability informed its design, including the choice of its neural architecture: RandLa-Net \cite{randlanet}.

Since the PointNet++ architecture \cite{pointnet2} succeeded to the ground-breaking PointNet \cite{pointnet} to operate directly on unordered point clouds, there were many attempts to improve over point-based architectures, characterized by PointNet-like operations hierarchically organized in a U-shaped architecture. Conceptually simple, RandLa-Net makes some interesting additions to PointNet++. It uses a lightweight module for local spatial encoding and achieves performance gains thanks to random sampling. Importantly, its authors demonstrated its segmentation accuracy on large-scale outdoor Lidar benchmark datasets like SemanticKITTI \cite{semantickitti_dataset} and Semantic 3D \cite{semantic3d_dataset}.

To produce the first experimental results on the benchmark, we kept the default hyperparameters of [3d-deep-learning-library]. Details about the processing of colorized Lidar HD tiles, about hyperparameters (optimizer, learning rate, scheduler, early stopping, etc.), and about infrastructure, are given in \cref{appendix:hparams}.

\subsection{Results}

\begin{table}
\caption{Baseline test IoUs and OA}\label{tab:ious}
\renewcommand{\arraystretch}{1}
\begin{tabularx}{\linewidth}{l>{\raggedleft\arraybackslash}X>{\raggedleft\arraybackslash}X}
\textbf{Class} & \textbf{IoU} & \textbf{OA} \\ \hline
other & 47.5 & 54.9 \\
ground & 91.9 & 97.7 \\
vegetation & 93.8 & 95.6 \\
building & 90.4 & 93.7 \\
water & 90.1 & 92.6 \\
bridge & 65.2 & 96.1 \\
permanent structure & 63.5 & 76.6 \\
\rowcolor{lightgray!60} \textbf{Macro Average} & 77.5 & 86.7 \\
\end{tabularx}
\end{table}

The baseline model achieves a test mIoU of 77.5\% and a test OA of 96.1\%. \cref{tab:ious} reports the IoU and OA for each class. We observe highly accurate results for the three most common classes: ground, vegetation, and building have an IoU above 90\%, with a maximum of 93.8\% for vegetation. The "water" class has an IoU of 90.1\%, despite its extreme initial rarity (0.6\%) in the area of interest. For the "bridge" and "permanent structure" classes, also initially rare (0.01\% each), the model achieves decent performance with an IoU above 60\%. The baseline underperforms for the "other" class compared to the six other ones, but still achieves an IoU of 47.5\%, which is remarkable considering the fuzzy definition of this class.

Additional metrics (precision, recall, and F1 score) are reported in Appendix \ref{appendix:metrics}. Confusions matrices are also given (Appendix \ref{appendix:confusion_matrices}). Finally, the qualitative performance of the baseline model can be assessed from examples of model predictions (Appendix \ref{appendix:samples}).

\section{Conclusion}
We present FRACTAL, an ultra-large-scale benchmark dataset for the 3D semantic segmentation of ALS point clouds. The dataset is based on high quality open data from the French Lidar HD acquisition program. It is a distillation of a larger initial area of 17,440 km² in five French regions. We show that a simple targeted sampling with spatial stratification at all levels preserves the diversity of regional-scale volumes of Lidar data. FRACTAL's size is compatible with deep learning research practices, and it is the largest Lidar benchmark dataset to date, with 9,261 million points in 100,000 point clouds and a total span of 250 km². While other ALS benchmark datasets typically cover a single urban area, the large diversity of urban and rural landscapes in FRACTAL is representative of the challenges of 3D semantic segmentation for land monitoring. 
Thanks to class rebalancing, our methodology opens the possibility of robust assessment of semantic segmentation performance, even for the initially rare classes water, bridge, and permanent structure. The baseline evaluation of a 3D neural network (RandLa-Net) further demonstrates the quality of the dataset for model evaluation. We invite the research community to benchmark both state-of-the-art and novel methods against this dataset. We hope that FRACTAL will advance the field of deep learning for airborne Lidar and ultimately benefit public Lidar-based 3D mapping programs for land monitoring.

\section*{Code and data access}
The dataset is made available on the HuggingFace platform as \href{https://huggingface.co/datasets/IGNF/FRACTAL}{IGNF/FRACTAL}.
The sampling, extraction, and colorization of point clouds were conducted with the Patch Catalog Sampling (PaCaSam) code repository available at: \href{https://github.com/IGNF/pacasam}{github.com/IGNF/pacasam}.
The baseline model was trained with the Myria3D code repository available at \href{https://github.com/IGNF/myria3d}{github.com/IGNF/myria3d}. Its weights are made available on HuggingFace as \href{https://huggingface.co/IGNF/FRACTAL-LidarHD_7cl_randlanet/}{IGNF/FRACTAL-LidarHD\_7cl\_randlanet}.

All assets are released under permissive open licences.

\section*{Acknowledgements}
The authors thank Léa Vauchier for her code reviews of the data engineering code, and Marouane Zellou for maintaining the in-house high-performance computing server, and Matthieu Porte and Anatol Garioud and Nicolas Audebert for reviewing this data paper. The authors also thank Anatol Garioud for paving the way for the open release of deep learning datasets for land monitoring at IGN with the FLAIR dataset \cite{flair_dataset}. 

\section*{Authors' contribution}
C.G. established the sampling framework. C.G. and F.R defined the data descriptors and their target proportions. C.G. implemented the tools for patch sampling and data extraction. F.R. and M.D. defined the specifications for the Lidar Patch Catalog, and M.D. implemented it. C.G. implemented and evaluated the deep learning baseline. C.G. wrote the data paper and released the dataset.

\clearpage
{\small
\bibliographystyle{ieee_fullname}
\bibliography{export}

\begin{thebibliography}{10}\itemsep=-1pt

\bibitem{semantickitti_dataset}
J. Behley, M. Garbade, A. Milioto, J. Quenzel, S. Behnke, C. Stachniss, and J. Gall.
\newblock {SemanticKITTI: A Dataset for Semantic Scene Understanding of LiDAR Sequences}.
\newblock In {\em Proc. of the IEEE/CVF International Conf.~on Computer Vision (ICCV)}, 2019.

\bibitem{landcoverai_dataset}
Adrian Boguszewski, Dominik Batorski, Natalia Ziemba-Jankowska, Tomasz Dziedzic, and Anna Zambrzycka.
\newblock Landcover.ai: Dataset for automatic mapping of buildings, woodlands, water and roads from aerial imagery.
\newblock In {\em Proceedings of the IEEE/CVF Conference on Computer Vision and Pattern Recognition (CVPR) Workshops}, pages 1102--1110, June 2021.

\bibitem{comet_ml}
{Comet ML}.
\newblock Comet [{ML} platform].
\newblock \url{www.comet.com}, 2024.

\bibitem{pytorch_lightning}
William Falcon and The PyTorch~Lightning team.
\newblock {PyTorch Lightning} [software].
\newblock \url{www.github.com/Lightning-AI/lightning}, 2019.

\bibitem{pytorch_geometric}
Matthias Fey and Jan~E Lenssen.
\newblock Fast graph representation learning with {PyTorch Geometric}.
\newblock In {\em ICLR Workshop on Representation Learning on Graphs and Manifolds}, 2019.

\bibitem{flair_dataset}
Anatol Garioud, Nicolas Gonthier, Loic Landrieu, Apolline De~Wit, Marion Valette, Marc Poup{\'e}e, S{\'e}bastien Giordano, et~al.
\newblock Flair: a country-scale land cover semantic segmentation dataset from multi-source optical imagery.
\newblock {\em Advances in Neural Information Processing Systems}, 36, 2024.

\bibitem{myria3d}
Charles Gaydon.
\newblock {Myria3D}: Deep learning for the semantic segmentation of aerial {Lidar} point clouds [software].
\newblock \url{www.github.com/IGNF/myria3d}, 2022.

\bibitem{semantic3d_dataset}
Timo Hackel, N Savinov, L Ladicky, Jan~D Wegner, K Schindler, and M Pollefeys.
\newblock {SEMANTIC3D.NET}: A new large-scale point cloud classification benchmark.
\newblock In {\em ISPRS Annals of the Photogrammetry, Remote Sensing and Spatial Information Sciences}, volume IV-1-W1, pages 91--98, 2017.

\bibitem{randlanet}
Qingyong Hu, Bo Yang, Linhai Xie, Stefano Rosa, Yulan Guo, Zhihua Wang, Niki Trigoni, and Andrew Markham.
\newblock {RandLa-Net}: Efficient semantic segmentation of large-scale point clouds.
\newblock In {\em Proceedings of the IEEE/CVF conference on computer vision and pattern recognition}, pages 11108--11117, 2020.

\bibitem{lidarhd}
{Institut national de l’information géographique et forestière (IGN)}.
\newblock {Lidar HD [Database]}.
\newblock \url{www.geoservices.ign.fr/documentation/donnees/alti/lidarhd}, 2023.

\bibitem{lidarhd_documentation}
{Institut national de l’information géographique et forestière (IGN)}.
\newblock {LiDAR HD} version 1.0 - descriptif de contenu des nuages de points {LiDAR}.
\newblock \url{www.geoservices.ign.fr/sites/default/files/2023-10/DC_LiDAR_HD_1-0_PTS.pdf}, 10 2023.

\bibitem{bdortho}
{Institut national de l’information géographique et forestière (IGN)}.
\newblock {ORTHO HR [Database]}.
\newblock \url{www.geoservices.ign.fr/bdortho}, 1 2023.

\bibitem{bdtopo}
{Institut national de l’information géographique et forestière (IGN)}.
\newblock {BD TOPO® [Database]}.
\newblock \url{www.geoservices.ign.fr/bdtopo}, 2024.

\bibitem{kakoulaki}
Georgia Kakoulaki, Ana Martinez, and Florio Petro.
\newblock Non-commercial {Light Detection and Ranging (LiDAR)} data in {Europe}.
\newblock Technical report, Joint Research Commission, 2021.

\bibitem{spatial_autocorrelation}
Teja Kattenborn, Felix Schiefer, Julian Frey, Hannes Feilhauer, Miguel~D. Mahecha, and Carsten~F. Dormann.
\newblock Spatially autocorrelated training and validation samples inflate performance assessment of convolutional neural networks.
\newblock {\em ISPRS Open Journal of Photogrammetry and Remote Sensing}, 5, 8 2022.

\bibitem{h3d_dataset}
Michael Kölle, Dominik Laupheimer, Stefan Schmohl, Norbert Haala, Franz Rottensteiner, Jan~Dirk Wegner, and Hugo Ledoux.
\newblock The {Hessigheim 3D (H3D)} benchmark on semantic segmentation of high-resolution {3D} point clouds and textured meshes from {UAV LiDAR and Multi-View-Stereo}.
\newblock {\em ISPRS Open Journal of Photogrammetry and Remote Sensing}, 1:100001, 10 2021.

\bibitem{wwf_lidar}
Markus Melin, Aurélie~C Shapiro, and Paul Glover-Kapfer.
\newblock {LiDAR} for ecology and conservation - {WWF} conservation technology series (3).
\newblock Technical report, WWF-UK, 2017.

\bibitem{lidar_flood_review}
Nur~Atirah Muhadi, Ahmad~Fikri Abdullah, Siti~Khairunniza Bejo, Muhammad~Razif Mahadi, and Ana Mijic.
\newblock The use of {LiDAR}-derived {DEM} in flood applications: a review.
\newblock {\em Remote Sensing}, 12, 7 2020.

\bibitem{vaihingen_isprs_dataset}
Joachim Niemeyer, Franz Rottensteiner, and Uwe Soergel.
\newblock Contextual classification of lidar data and building object detection in urban areas.
\newblock {\em ISPRS Journal of Photogrammetry and Remote Sensing}, 87:152--165, 1 2014.

\bibitem{pointnet}
Charles~R Qi, Hao Su, Kaichun Mo, and Leonidas~J Guibas.
\newblock Pointnet: Deep learning on point sets for {3D} classification and segmentation.
\newblock In {\em Proceedings of the IEEE conference on computer vision and pattern recognition}, pages 652--660, 2017.

\bibitem{pointnet2}
Charles~Ruizhongtai Qi, Li Yi, Hao Su, and Leonidas~J Guibas.
\newblock {Pointnet++}: Deep hierarchical feature learning on point sets in a metric space.
\newblock {\em Advances in neural information processing systems}, 30, 2017.

\bibitem{opengf_dataset}
Nannan Qin, Weikai Tan, Lingfei Ma, Dedong Zhang, and Jonathan Li.
\newblock {OpenGF}: An ultra-large-scale ground filtering dataset built upon open {ALS} point clouds around the world.
\newblock In {\em Proceedings of the IEEE/CVF conference on computer vision and pattern recognition}, pages 1082--1091, 2021.

\bibitem{asprs_las}
{The American Society for Photogrammetry and Remote Sensing}.
\newblock {LAS} specification 1.4 - {R15}.
\newblock Technical report, The American Society for Photogrammetry and Remote Sensing, 2019.

\bibitem{dales_dataset}
Nina Varney, Vijayan~K Asari, and Quinn Graehling.
\newblock {DALES}: A large-scale aerial {LiDAR} data set for semantic segmentation.
\newblock In {\em Proceedings of the IEEE/CVF conference on computer vision and pattern recognition workshops}, pages 186--187, 2020.

\bibitem{lasdu_dataset}
Zhen Ye, Yusheng Xu, Rong Huang, Xiaohua Tong, Xin Li, Xiangfeng Liu, Kuifeng Luan, Ludwig Hoegner, and Uwe Stilla.
\newblock {LASDU}: A large-scale aerial {LiDAR} dataset for semantic labeling in dense urban areas.
\newblock {\em ISPRS International Journal of Geo-Information}, 9, 7 2020.

\bibitem{fire_lidar_review}
M Yebra, S Marselis, A van Dijk, G Cary, and Y Chen.
\newblock Using {Lidar} for forest and fuel structure mapping: Options, benefits, requirements and costs.
\newblock Technical report, Bushfire and Natural Hazards CRC, 2015.

\bibitem{lidar_software_review}
Kin Yen.
\newblock Automated {LiDAR} extraction software.
\newblock Technical report, Caltran's Division of Research, Innovation and System Information, 2021.

\bibitem{cenagis_als_dataset}
P. Zachar, K. Bakuła, and W. Ostrowski.
\newblock {CENAGIS-ALS Benchmark} - new proposal for dense {ALS} benchmark based on the review of datasets and benchmarks for {3D} point cloud segmentation.
\newblock volume~48, pages 227--234. International Society for Photogrammetry and Remote Sensing, 10 2023.

\bibitem{dublincity_dataset}
S.~M.~Iman Zolanvari, Susana Ruano, Aakanksha Rana, Alan Cummins, Rogerio~Eduardo da Silva, Morteza Rahbar, and Aljosa Smolic.
\newblock {DublinCity}: Annotated {LiDAR} point cloud and its applications.
\newblock 9 2019.

\end{thebibliography}
}

\onecolumn\clearpage
\appendix
\section*{Appendices of the FRACTAL datapaper}
\captionsetup{labelformat=empty,justification=raggedright,singlelinecheck=false, textfont=it}

\FloatBarrier

\section{Data representation for patch sampling}
\subsection{Considerations for data cataloguing}
We present the data representation used for sampling, which takes the form of a PostGIS database. Cataloging data boils down to two steps: listing and describing, which we illustrate in \cref{fig:lipac}. 
We start by regularly dividing the area of interest into $50 \times 50$~m~patches. We then describe all patches with two sets of descriptors: Lidar descriptors and vector descriptors. Lidar descriptors are derived from the Lidar data patches themselves: number of points in each semantic class, elevation (above sea level), and elevation gain. Vector descriptors are derived by cross-referencing the patches with the BD TOPO® \cite{bdtopo}, a 3D vector description of the French territory and its infrastructures (buildings, roads, etc.). We include flags indicating the presence of the following objects: greenhouses, antennas, pylons, highways, parking lots, water surfaces, greenhouses, pylons, antennas, and industrial, commercial, or agricultural buildings.

Cataloguing with a PostGIS database gives us the full power of SQL syntax to select data and to further characterize each Lidar patch with bespoke SQL queries that combine simple descriptors into more complex ones (see \cref{tab:descriptors}).

\subsection{Data cataloguing}\label{fig:lipac}
\begin{figure*}[h!]
 \centering
    \includegraphics[width=1.0\textwidth, alt={The figure shows how a single Lidar HD Tile of 1 km x 1 km is divided into many smaller patches which are then described with vectore features and point cloud features.}]{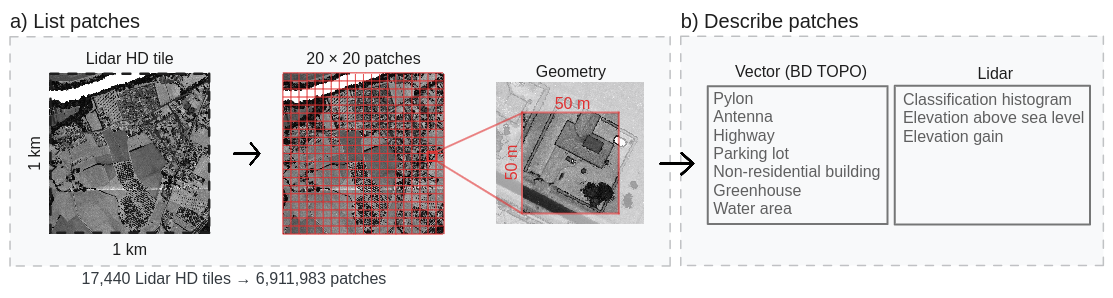}
    \caption{Data cataloguing boils down to a) listing and b) describing all Lidar patches, prior to sampling.}
\end{figure*}

\FloatBarrier\clearpage
\section{Dataset content}
\subsection{Number of points by class in the area of interest, in the dataset, and in the train, val and test sets}\label{appendix:all_sizes}
\begin{table}[h!]
\setlength{\tabcolsep}{2.2pt}
\renewcommand{\arraystretch}{1}
\begin{tabular}{p{3.5cm}cc|cc|cc|cc|cc}
\multicolumn{3}{c}{} & \multicolumn{8}{c}{\textbf{FRACTAL}}  \\ \cline{4-11}
 & \multicolumn{2}{c}{\textbf{Area of Interest}} & \multicolumn{2}{c}{\textbf{FRACTAL}} & \multicolumn{2}{c}{\textbf{Train}} & \multicolumn{2}{c}{\textbf{Val}} & \multicolumn{2}{c}{\textbf{Test}}   \\ \cline{2-3}\cline{4-5}\cline{6-7}\cline{8-9}\cline{10-11}
\textbf{Name} & \textbf{Points (M)} & \textbf{\%} & \textbf{Points (M)} & \textbf{\%} & \textbf{Points (M)} & \textbf{\%} &\textbf{ Points (M)} & \textbf{\%} & \textbf{Points (M)} & \textbf{\%}   \\ \hline
\textbf{Other} & 2,137 & 0.32 & 53 & 0.57 & 41 & 0.56 & 5 & 0.53 & 6 & 0.66  \\
\textbf{Ground} & 258,751 & 39.1 & 3,625 & 39.1 & 2,874 & 39.0 & 359 & 39.1 & 391 & 40.5  \\
\textbf{Vegetation} & 394,425 & 59.6 & 5,248 & 56.7 & 4,203 & 57.0 & 523 & 56.9 & 523 & 54.1  \\
\textbf{Building} & 4,939 & 0.75 & 264 & 2.85 & 206 & 2.80 & 26 & 2.80 & 32 & 3.33  \\
\textbf{Water} & 1,504 & 0.23 & 54 & 0.59 & 38 & 0.52 & 5 & 0.49 & 12 & 1.20  \\
\textbf{Bridge} & 47 & 0.01 & 12 & 0.13 & 9 & 0.13 & 1 & 0.10 & 2 & 0.16  \\
\textbf{Permanent structure} & 33 & 0.01 & 3 & 0.04 & 3 & 0.04 & 0 & 0.04 & 0 & 0.03  \\
\rowcolor{lightgray!60} \textbf{Total} & 661998 & - & 9261 & - & 7376 & - & 919 & - & 966 & -  \\
\end{tabular}
\end{table}

\subsection{Proportions of scene types in the train, val, and test sets}\label{appendix:descriptors_trainvaltest}
\begin{table}[h!]
\setlength{\tabcolsep}{4.9pt}
\renewcommand{\arraystretch}{1}
\begin{tabular}{lccc}
\textbf{Descriptor} & \textbf{Train (\%)} & \textbf{Val (\%)} & \textbf{Test (\%)}  \\ \hline \arrayrulecolor{lightgray!60}
\texttt{BUILD} & 22.5 & 22.8 & 25.7  \\
\texttt{BUILD\_BIG} & 5 & 4.8 & 4.9  \\
\texttt{BUILD\_GREENHOUSE} & 1.3 & 1.3 & 1.2  \\ \hline 
\texttt{BRIDGE} & 5.5 & 5.4 & 5.4  \\ \hline 
\texttt{WATER} & 8.2 & 8.1 & 11  \\ 
\texttt{WATER\_SURFACE} & 6 & 5.9 & 7.7  \\ \hline 
\texttt{PERMSTRUCT} & 5 & 4.9 & 4.6  \\
\texttt{PERMSTRUCT\_ANTENNA} & 0.9 & 0.9 & 0.6  \\
\texttt{PERMSTRUCT\_PYLON} & 1.5 & 1.4 & 1.4  \\ \hline 
\texttt{OTHER} & 22.7 & 22.6 & 27.5  \\
\texttt{OTHER\_HIGHWAY} & 1.3 & 1.2 & 1.7  \\
\texttt{OTHER\_PARKING} & 1.1 & 1.1 & 1.5  \\ \hline 
\texttt{FOREST} & 7.3 & 7.4 & 7.1  \\
\texttt{HIGHSLOPE1} & 11.5 & 11.8 & 11  \\
\texttt{HIGHSLOPE2} & 5.5 & 5.4 & 4.8  \\ 
\texttt{MOUTAIN} & 17.4 & 17.6 & 14.5  \\ 
\texttt{WATER\_ONLY} & 1.7 & 1.7 & 4.5  \\
\texttt{SEASHORE} & 1.4 & 1.4 & 4.5  \\ 
\texttt{URBAN} & 6.4 & 6.5 & 7.4  \\
\end{tabular}
\end{table}

\begin{figure*}
    \subsection{Patches in FRACTAL on a subset area of 4 × 6 Lidar HD tiles}\label{appendix:sampling_qgis_4x6}
    \vspace{-0.2cm}
    \centering
    \includegraphics[width=0.9\textwidth]{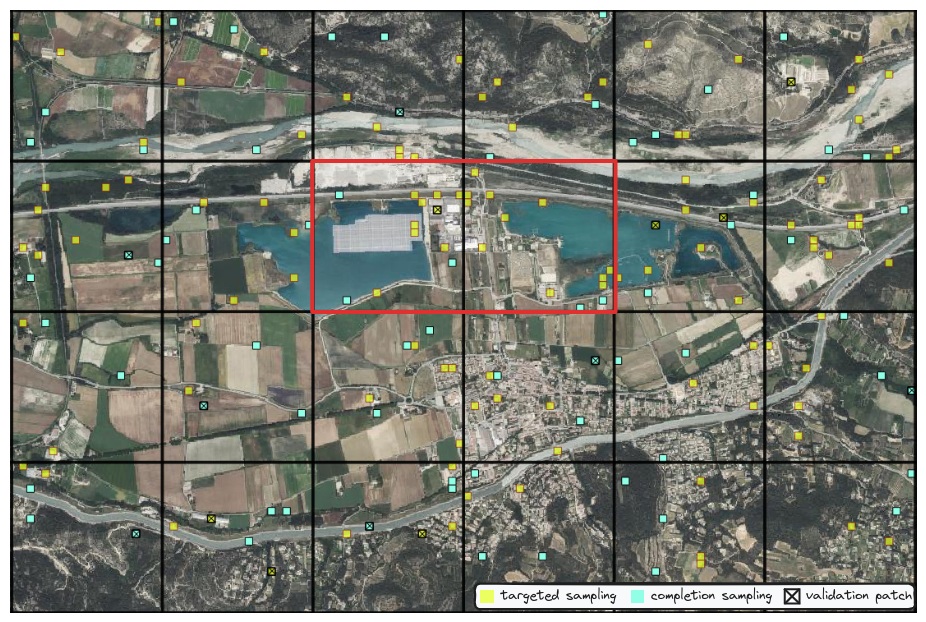}
    \caption{The 182 selected patches cover only 1.9\% of the 24 km² (52-fold reduction). Patches selected via targeted sampling (yellow) are more present in complex urban areas. Patches from completion (cyan) sampling are homogeneously distributed. Validation patches (crossed in black) are homogeneously distributed across Lidar HD tiles and scene types. The two tiles framed in red are the ones displayed in \cref{fig:sampling_qgis_1x2}}
\end{figure*}

\FloatBarrier
\clearpage

\section{Baseline model evaluation}\label{appendix:modelevaluation}

\subsection{Baseline test IoU, precision, recall, and F1 score, for each semantic class and macro-averaged}\label{appendix:metrics}
\begin{table}[h!]
\setlength{\tabcolsep}{4.9pt}
\renewcommand{\arraystretch}{1}
\begin{tabular}{lccccc}
\textbf{Class} & \textbf{IoU} & \textbf{Accuracy} & \textbf{Precision} & \textbf{Recall} & F1 Score  \\ \hline
other & 47.5 & 54.9 & 77.8 & 54.9 & 64.4  \\
ground & 91.9 & 97.7 & 93.8 & 97.7 & 95.8  \\
vegetation & 93.8 & 95.6 & 98.0 & 95.6 & 96.8  \\
building & 90.4 & 93.7 & 96.2 & 93.7 & 95.0  \\
water & 90.1 & 92.6 & 97.1 & 92.6 & 94.8  \\
bridge & 65.2 & 96.1 & 79.3 & 78.6 & 79.0  \\
permanent structure & 63.5 & 76.6 & 78.9 & 76.6 & 77.7  \\
\rowcolor{lightgray!60} \textbf{Macro Average} & 77.5 & 86.7 & 88.7 & 84.2 & 86.2  \\

\end{tabular}
\end{table}

\subsection{Baseline test confusion matrices normalized by rows (a) and columns (b)}\label{appendix:confusion_matrices}
\begin{figure*}[h!]
 \centering
    \includegraphics[width=1.0\textwidth]{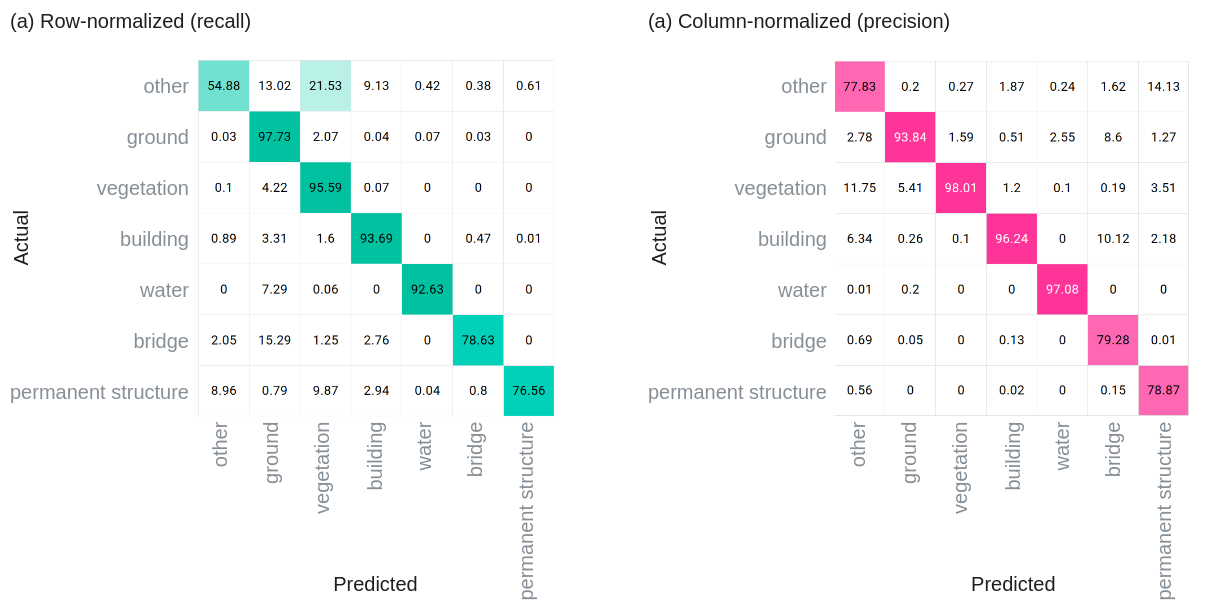}
    \caption{Diagonal values in row-normalized matrix correspond to recall. Diagonal values in column-normalized matrix correspond to precision.}

    \caption{Vegetation, ground, and building have low relative confusion, as expected from their high IoU. Permanent structure has its highest confusion with vegetation, due to the vertical nature and intricate geometries shared by both classes e.g., between pylons and trees. High interclass confusion is also found between classes that may have imprecise geometric boundaries. For example, bridge points are misclassified as ground, of which they are often an extension. Points of class other have high confusion with vegetation and ground points. This is mainly due to the Lidar HD classification specification, which favors specificity over recall for the ground and vegetation classes. As a result, the model may be penalized for accurately identifying ground and vegetation whose target class is other. Note that with only 0.6\% of all points in FRACTAL, class other is a small minority, and these errors have a negligible effect on the measured performance for the ground and vegetation classes.}
 
\end{figure*}

\subsection{Input cloud, target classification, and prediction of baseline model for a random subset of patches}\label{appendix:samples}
\begin{figure*}[h!]
 \centering
    \includegraphics[width=0.9\textwidth]{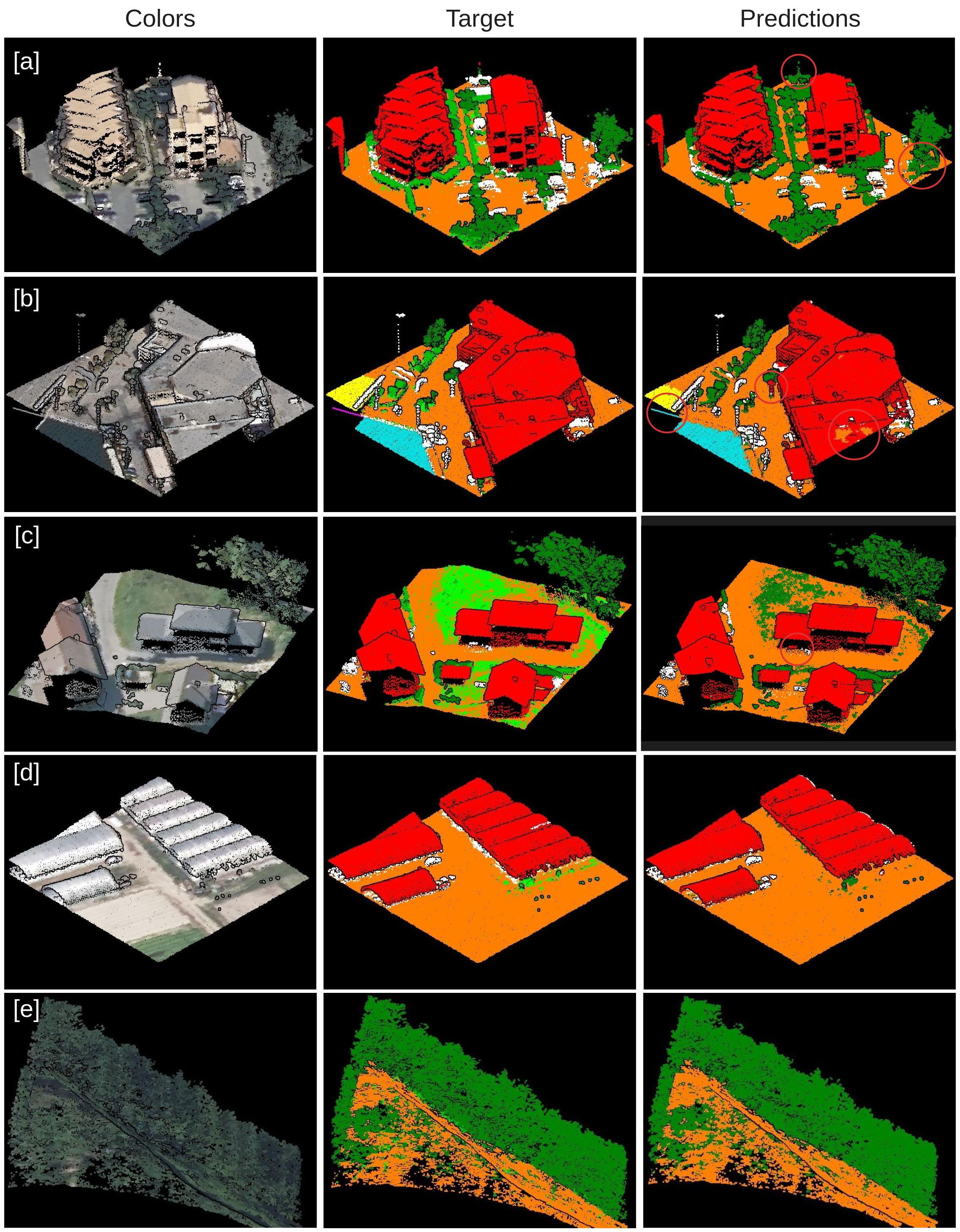}
    \caption{Test patches are selected at random and without cherry-picking, among test patches with at least 10k points (at least 4 pts/m²), to match the following scene types: a) \texttt{OTHER\_PARKING}, b) \texttt{WATER} and \texttt{BRIDGE}, c) \texttt{URBAN}, d) \texttt{BUILD\_GREENHOUSE}, and e) \texttt{HIGHSLOPE1}. Color scheme is: other (white), ground (orange), vegetation (green), building (red), water (cyan), bridge (yellow), permanent structure (purple). Predictions errors are circled in red.}
\end{figure*}

\FloatBarrier
\subsection{Preprocessing and training hyperparameters}\label{appendix:hparams}

Lidar is unstructured; how we feed clouds to a model is important, and has a lot of degrees of freedom. We first subsample clouds via a grid with a voxel size of 0.25 m. A maximal budget of 40,000 points per cloud is allocated, enforced when needed via a random subsampling. Clouds are horizontally centered by subtracting their average position along the x and y axes, and vertically aligned by subtracting their minimal position along the z axis. Centered coordinates are then divided by 25 (meters) to be homogeneously brought between –1 and 1. We apply several generic data augmentations, namely: Random Scale (x0.8 to x1.25 factor), Random Jitter (-0.05 m to +0.05 m along each axis), Random Translate (-1 m to +1 m), Random Flip (along x and y axes).

In terms of features, the following dimensions are included: x, y, z, reflectance, echo number, number of echos. Number of echos and echo number are normalized by constants to be between 0 and 1. Reflectance is log-normalized, standardized, and values above three standard deviations are clamped.
The point cloud is colorized, and we also consider the near infrared, red, green, blue dimensions. We set all colors to 0 for points with an echo number above 1, as a very basic occlusion model. Furthermore, we compute the Normalized Difference Vegetation Index (NDVI). We also calculate a new “color intensity” feature as the average of red, green, and blue channels. Colors, color intensity, and NDVI, are divided by a constant to be between 0 and 1. Average color is normalized similarly to the reflectance, i.e., by log-normalization, standardization, and truncation of values above three standard deviations.

Training is supervised with Cross-Entropy loss and the Adam optimizer. The learning rate is 0.004, with a reduction strategy (ReduceLROnPlateau) that halves the learning rate with a patience of 20 epochs and a cooldown of 5 epochs after each reduction. A batch size of 10 is used. We train the model for at least 100 epochs and retain the model that minimizes the validation loss.

Training is conducted with 6 NVIDIA Tesla V100 GPUs, each equipped with 32 GB of memory, using Pytorch-Lightning's Distributed Data Parallel (DDP) strategy. The approximate learning time is 30 minutes per epoch. We log metrics using Comet, a machine learning experiment tracking tool \cite{comet_ml}.

\FloatBarrier\clearpage
\section{Data sources specifications}

\subsection{Specifications of the Lidar HD classification}\label{appendix:lhd_nomenclature}
\begin{table}[h!]
\setlength{\tabcolsep}{4.9pt}
\renewcommand{\arraystretch}{1.}
\begin{tabular*}{\linewidth}{lccp{0.6\linewidth}}
\textbf{RGB}  & \textbf{Value}      & \textbf{Name}   & \textbf{Content} \\ \hline
\cellcolor[RGB]{255,255,255}255,255,255 & 1 & Unclassified &  All points that do not belong in any of the other classes. For example, it includes vehicles, animals or people, temporary objects, wood piles, etc.   \\ \hline
\cellcolor[RGB]{255,128,0} 255,128,0 & 2 & Ground & Points located on the surface of natural and artificial ground. Bridge decks are not part of this class. \\ \hline
\cellcolor[RGB]{0,255,0} 0,255,0 & 3 & Low vegetation & \multirow{3}{\linewidth}{Trees, shrubs and low vegetation (e.g. bushes, ferns, reeds, etc.). Vegetation at ground level (less than 20 cm high, typically grass) is classified as vegetation only if there are enough ground points locally for ground modeling. The class also includes cultivated trees (orchards, vineyards, field crops, etc.). Vegetation is further divided into 3 classes based on height above ground: low (< 0.5m), medium (between 0.5 m and 1.5 m), and high ($\geq{1.5~m}$) vegetation.} \\[16pt]
\cellcolor[RGB]{0,193,0} 0,193,0 & 4 & Medium vegetation &   \\[16pt]
\cellcolor[RGB]{0,134,0} 0,134,0 & 5 & High vegetation &   \\[16pt] \hline
\cellcolor[RGB]{255,0,0} 255,0,0 & 6 & Building & A building is defined as a permanent structure with an area greater than 10~m². Besides residential buildings it includes monuments, castles, mills, water towers, lighthouses, industrial chimneys, ramparts and fortifications. The specification also includes roofs and façades, as well as chimneys, dormer windows, skylights and balconies. As the permanent nature of a structure cannot be established from lidar data only, this class may include lightweight structures without walls such as garden sheds, bungalows, market canvases or awnings. \\ \hline
\cellcolor[RGB]{0,225,225} 0,225,225 & 9 & Water & All points located on the surface of rivers, bodies of water, sea, or ocean.   \\ \hline
\cellcolor[RGB]{255,255,0} 255,255,0 & 17 & Bridge deck & A bridge is an engineering structure passing over one or more elements of the road, rail or waterway network. Point on bridge decks are included, while structural elements such as piers and parapets are assigned to the "Unclassified" class. Very high structural elements ($\geq$ 5 m above deck level) such as piers and abutments are classified as “Permanent structures”. Tunneled passages (including nozzles, which are openings in the ground generally to allow water to drain) are considered part of the ground and therefore excluded.  \\ \hline
\cellcolor[RGB]{128,0,64} 128,0,64 & 64 & Permanent structures & All aboveground objects other than buildings, vegetation and bridges, that are identified as perennial and of such a nature that they characterize the landscape. This class includes (but is not limited to): wind turbines, cable cars, telecommunication antennas, electricity distribution networks (cables and pylons), bridge elements above the deck (cables, piers, etc.).  \\ \hline
\cellcolor[RGB]{64,0,128} 64,0,128 & 65 & Artifact & All points that do not correspond to an actual object or terrain. \\ \hline
\cellcolor[RGB]{255,0,255} 255,0,255 & 66 & Synthetic &  Artificial points created under bridges and and on water surfaces to have coherent digital models.  \\ \hline

\end{tabular*}
\caption{Refer to the Lidar HD product description \cite{lidarhd_documentation} for more specifications.}
\end{table}\FloatBarrier\clearpage

\subsection{Relation between the Lidar HD classification nomenclature and its adaptation in FRACTAL}\label{appendix:nomenclatures}
\begin{table}
\setlength{\tabcolsep}{4.9pt}
\renewcommand{\arraystretch}{1}
\begin{tabular}{ccc}
\textbf{Lidar HD}   & $\rightarrow$ & \textbf{FRACTAL} \\ \hline \arrayrulecolor{lightgray!60}
Unclassified & $\rightarrow$ & Other   \\ \hline
Ground & $\rightarrow$ & Ground  \\ \hline
Low vegetation &  & \multirow{3}{*}{Vegetation}  \\
Medium vegetation & $\rightarrow$ &   \\
High vegetation & &   \\ \hline
Buildings & $\rightarrow$ & Building  \\ \hline
Water & $\rightarrow$ & Water  \\ \hline
Bridge deck & $\rightarrow$ & Bridge  \\ \hline
Permanent structures & $\rightarrow$ & Permanent structure  \\ \hline
Artifact &  & \multirow{2}{*}{-} \\
Synthetic &  &   \\ \hline
\end{tabular}
\end{table}

\begin{table}[h!]
\subsection{Sensors used to acquire Lidar HD data in each of the 8 50 $\times$ 50 km blocks represented in FRACTAL}\label{appendix:sensors_subcontractors}
\setlength{\tabcolsep}{4.9pt}
\renewcommand{\arraystretch}{1}
\begin{tabular}{lccc}
 &  & \multicolumn{2}{c}{\textbf{Subcontractor(s)}} \\ \cline{3-4}
\textbf{Block} & \textbf{Sensor} & \textbf{Acquisition} & \textbf{Classification}  \\ \hline
GN & CityMapper 2H: Hyperion 2+  & APEI + Avineon & Avineon  \\
MQ & RIEGL VQ1560 II-S  & Eurosense + SFS & Eurosense + SFS  \\
MP & RIEGL VQ-1560 II  & Eurosense & Sintégra  \\
PK & RIEGL VQ1560 II  & Eurosense & Eurosense  \\
PO & Leica CityMapper-2  & Avineon + APEI & Avineon + APEI  \\
PP & Leica CityMapper-2  & Avineon + APEI & Eurosense + SFS  \\
QO & RIEGL VQ780 II-S  & Sintégra + Bluesky & Sintégra + Bluesky  \\
UT & RIEGL VQ780 II-S  & Sintégra + Bluesky & Avineon + APEI  \\ \hline
\end{tabular}
\caption{Data acquisition and its classification were often conducted by distinct subcontractors, which we also report.}
\end{table}

\FloatBarrier
\section{Structure of files and directories}\label{appendix:dataset_structure}

\begin{figure*}[h!]
    \centering
    \renewcommand{\arraystretch}{1.5}
    \par\vskip1.2pt
    \hspace{0.02cm}
    \begin{Tabular}[0.8]{|p{8.2cm}|}
    \hline 
    \rowcolor[HTML]{f7f9fd}  {\footnotesize
    \begin{forest}
      pic dir tree, where level=0{}{directory,},
      for tree={ s sep=0.05cm, l sep=0.65cm, font=\rmfamily }
      [\textbf{FRACTAL}
        [\textbf{data}
          [\textbf{train}
              [\textbf{00 ... 79}
                [\texttt{TRAIN-\{tile\_id\}-\{patch\_id\}}.laz, file]
              ]
          ]
          [\textbf{val}
              [\textbf{00 ... 09}
                [\texttt{VAL-\{tile\_id\}-\{patch\_id\}}.laz, file]
              ]
          ]
        [\textbf{test}
              [\textbf{00 ... 09}
                [\texttt{TEST-\{tile\_id\}-\{patch\_id\}}.laz, file]
              ]
          ]
        ]
        [fractal.gpkg, file]
      ]
    \end{forest}}      \\ \hline
    \end{Tabular}
    
\caption{The point clouds are organized by set (train, val, and test), in 100 subdirectories of 1000 point clouds each. The naming convention defines \texttt{tile\_id} as the X and Y kilometer northwest coordinates of the Lidar HD tile in the Lambert 93 projection (EPSG:2154), and \texttt{patch\_id} is a unique, arbitrary patch identifier. For instance, patch \texttt{TEST-0744\_6246-006804306.laz} belongs to the test set, was extracted from Lidar HD tile whose top left coordinates are $X=0744$ and $Y=6246$, and its unique identifier is $006804306$.}
\caption{The geometries of all patches and their descriptions are provided in a metadata file: \texttt{fractal.gpkg}.}
\end{figure*}

\end{document}